\documentclass[journal]{IEEEtran}
\usepackage{graphicx,color,psfrag}
\usepackage{amsmath}
\usepackage{amssymb}
\usepackage{bm}

\usepackage{algorithmic} 
\usepackage{algorithm} 
\usepackage{url}

\begin{document}
%
\bstctlcite{IEEEexample:BSTcontrol}
\title{Autonomous Navigation of Micro Air Vehicles in Warehouses Using Vision-based Line Following}
%
%
%
\author{Ling~Shuang~Soh,~and~Hann~Woei~Ho*%
\thanks{*Corresponding author}%
\thanks{Ling Shuang Soh and Hann Woei Ho are with School of Aerospace Engineering, Engineering Campus, Universiti Sains Malaysia, 14300 Nibong Tebal, Pulau Pinang, Malaysia (email: aehannwoei@usm.my)}%
}

\markboth{}%
{Shell \MakeLowercase{\textit{et al.}}: }
%



\maketitle

\begin{abstract}

In this paper, we propose a vision-based solution for indoor Micro Air Vehicle (MAV) navigation, with a primary focus on its application within autonomous warehouses. Our work centers on the utilization of a single camera as the primary sensor for tasks such as detection, localization, and path planning. To achieve these objectives, we implement the HSV color detection and the Hough Line Transform for effective line detection within warehouse environments. The integration of a Kalman filter into our system enables the camera to track yellow lines reliably. We evaluated the performance of our vision-based line following algorithm through various MAV flight tests conducted in the Gazebo 11 platform, utilizing ROS Noetic. The results of these simulations demonstrate the system capability to successfully navigate narrow indoor spaces. Our proposed system has the potential to significantly reduce labor costs and enhance overall productivity in warehouse operations. This work contributes to the growing field of MAV applications in autonomous warehouses, addressing the need for efficient logistics and supply chain solutions.

\end{abstract}


%
\IEEEpeerreviewmaketitle

\section{Introduction}
\label{sec:intro}

Micro air vehicles (MAVs) are unmanned aerial vehicles (UAVs) characterized by their small size, lightweight, and agility. They have gained popularity in military and civilian applications, including aerial photography, recreational use, and competitive drone racing. MAVs can navigate tight spaces and capture high-resolution images or videos from unique perspectives. They have also gained importance as a valuable tool for victim search and rescue, surveillance operations, scientific research endeavors, and reconnaissance missions \cite{cai2010brief}. 

Traditionally, warehouse tasks, such as goods reception, storage, picking, and dispatch have relied on manual labor by human operators. However, this approach is often time-consuming, labor-intensive, and prone to errors. The concept of autonomous warehouses has gained significant traction as a transformative solution for improving efficiency and productivity in the logistics and supply chain industry. Autonomous warehouses are characterized by the integration of advanced technologies, such as robotics, artificial intelligence (AI), and automation systems, to enable self-operating and self-optimizing capabilities \cite{da2021robotic}. Systems include autonomous robots, automated guided vehicles (AGVs), and conveyor systems that work collaboratively to handle inventory movement and management \cite{tran2021physical}. 

The use of MAVs in warehouse environments has garnered significant attention as a potential solution for enhancing efficiency and optimizing operations. One specific application that has emerged is the development of line-following MAVs in warehouses \cite{anand2019grid}. These MAVs are specifically designed to navigate predetermined paths or lines on the warehouse floor to perform tasks such as inventory management, item tracking, and monitoring. The introduction of line-following MAVs also enables the automation of tasks, resulting in enhanced accuracy, reduced labor costs, and improved overall productivity in warehouse operations.

Line-following MAVs typically employ a combination of sensors, onboard cameras, and intelligent algorithms to detect and track lines or paths on the warehouse floor. Computer vision techniques are utilized to analyze visual information captured by the onboard cameras, enabling the identification of the line or path. The MAVs then adjust their flight trajectory to maintain alignment with the designated path, ensuring accurate navigation.

\section{Related Works}
\label{sec:related}
Research on agile flight, including vision-based localization and control algorithms, has been conducted through a combination of experimental and computational approaches. Experimental investigations typically involve real-world environments that are scaled down and equipped with various sensors to capture relevant data \cite{li2020autonomous}. Researchers use these setups to evaluate the performance of agile flight algorithms and assess their effectiveness in localization and control.

Indoor localization technology refers to the methods and techniques used to determine the precise location of objects or individuals within indoor environments where GPS signals may not be available or reliable. There are various indoor localization techniques, including angle of arrival (AoA), time of flight (ToF), return time of flight (RTOF), and received signal strength (RSS). It covers a wide range of technologies utilized in indoor localization, such as Wireless Fidelity (Wi-Fi), RFID, ultra-wideband (UWB), and Bluetooth \cite{zafari2019survey, ching2022ultra}. 

One of the techniques, the RSS-based approach is a popular and straightforward method employed for indoor localization. It relies on measuring the strength of the signal received at the receiver. By utilizing RSS, it becomes possible to get the distance between a transmitter (Tx) and a receiver (Rx), providing valuable information for indoor localization purposes \cite{zafari2019survey}. The signal strength received by the receiver exhibits variations over time, particularly in relation to the relative position of the transmitter and receiver \cite{ladd2002robotics}. However, the RSS-based approach encounters significant performance degradation in complex scenarios characterized by factors such as multipath fading and temporal dynamics \cite{yang2013rssi}. 

Other than the techniques and technologies mentioned earlier, vision-based localization is also one of the methods for indoor localization. Vision-based localization is a technique that relies on visual information captured by cameras to estimate the position and orientation of a device within an indoor environment. It involves analyzing the visual data to extract relevant features or landmarks, such as objects, patterns, or markers, which can be used as reference points for localization. Vision-based solutions do provide the advantages of portability, compactness, cost-effectiveness, and power efficiency \cite{lee2011vision}. The proposed method involves extracting points from images captured by a camera using the Scale Invariant Feature Transform (SIFT) algorithm. Landmark feature points with distinctive descriptor vectors are selected and their locations are detected and stored in a map database to estimate the position of the UAV. Furthermore, the utilization of Speed-Up Robust Features (SURF) techniques in vision-based localization helps decrease the storage overhead of the database, resulting in reduced latency in real-world applications \cite{guan2016vision}.

Furthermore, more and more autonomous warehouses are gradually replacing traditional warehouses to improve efficiency and productivity in the logistics and supply chain industry, also indirectly reducing labor costs \cite{poudel2013coordinating}. Autonomous warehouses are characterized by the integration of advanced technologies, such as robotics, AI, and automation systems, to enable self-operating and self-optimizing capabilities \cite{da2021robotic}. Systems include autonomous robots, AGVs, and conveyor systems that work collaboratively to handle inventory movement and management \cite{tran2021physical}. One of the classic modes of autonomous warehouse is the cooperation between Unmanned Ground Vehicle (UGV) and UAV. Since the warehouse inventory is conducted indoors where GPS cannot be used, the UGV is employed as a ground reference for the UAV to overcome expensive and impractical indoor localization techniques \cite{guerin2016towards}.

Integrated systems like EyeSee and Infinium Scan have been developed as Unmanned Aerial Systems (UAS) platforms tailored for warehouse environments and the purpose of inventory tracking \cite{barlow2019multi}. One notable example is the world's foremost furniture manufacturer and retail chain, IKEA had employs drones to perform inventory checks.  IKEA has implemented drones in five of its warehouse stores and one distribution center, starting in December 2021. In the IKEA Malmo and Padua setups, prototype EyeSee drones from Hardis were utilized, which is particularly suitable for distribution centers that require less frequent inventory counting and where inventory accuracy is generally of lesser importance \cite{maghazei2022emerging}. 

\begin{figure}[ht]
\centering
\includegraphics[trim=0 0 0 0, clip, width =0.3\textwidth]{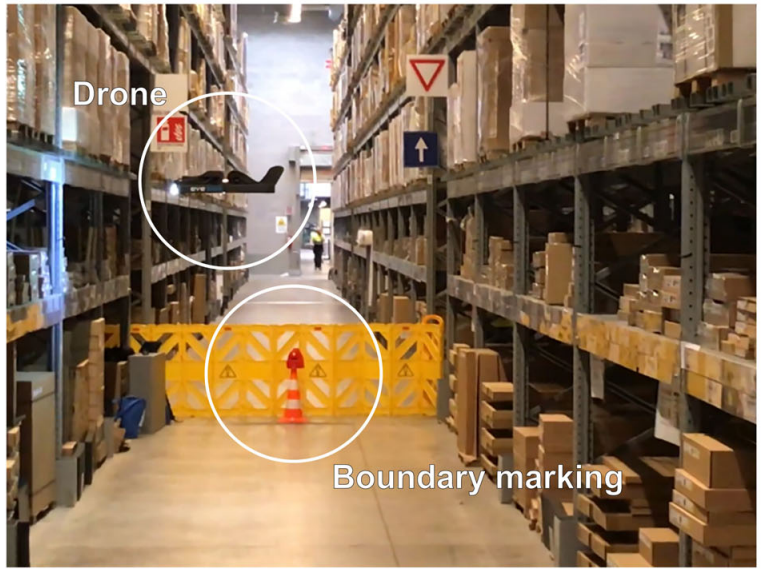}
\caption{Semi-autonomous drone “EyeSee” used in IKEA Padua warehouse. \cite{maghazei2022emerging}}
\label{fig:eyesee}
\end{figure}

Another prominent example of autonomous robot systems is the most famous one, Amazon robotics solution, which has 520,000 mobile drive unit robots worldwide. Various AGVs and Autonomous Mobile Robots (AMRs) such as Pegasus, Kiva, Hercules, and Proteus are utilized in these systems. Warehouse mapping, object detection, and navigation in Amazon warehouses rely on the utilization of RFID scanners and Light Detection and Ranging (LiDAR) \cite{IREMOS27121}. The study also showed that by implementing parallel execution of the placing operation and vision recognition algorithms, the mission time was reduced from 42 seconds to 31 seconds compared to human work \cite{del2017uji}. 

\begin{figure}[ht]
\centering
\includegraphics[trim=0 0 0 0, clip, width =0.3\textwidth]{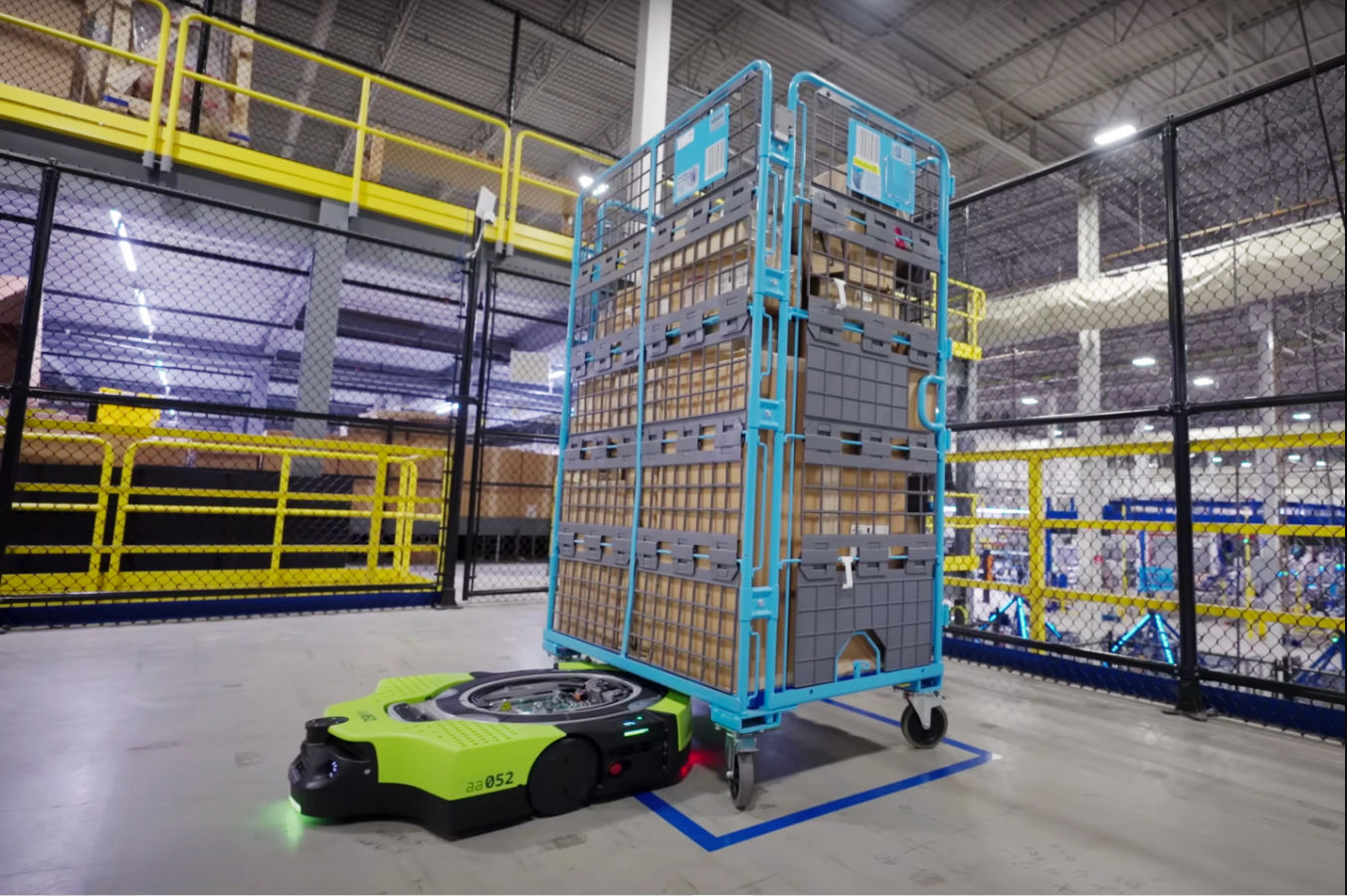}
\caption{Shelving units carried by an Amazon Proteus. \cite{IREMOS27121}}
\label{fig:proteus}
\end{figure}

The control aspect of agile flight presents significant challenges due to the high speed at which MAVs operate. The fast speed introduces nonlinear dynamics, complex aerodynamic effects, and constraints on the actuators that control the vehicle's motion \cite{sun2022comparative}. To address these challenges, researchers have explored different control strategies and compared their performance \cite{song2023reaching, zhou2017launch}.

In a study, the authors compared the performance of two control methods: Nonlinear Model Predictive Control (NMPC) and Differential-flatness-based controller (DFBC) \cite{sun2022comparative}. Flight tests were conducted in both real-world scenarios and simulations. The results indicated that the NMPC effectively handled infeasible trajectories through future prediction. Additionally, when combined with the Incremental Nonlinear Dynamic Inversion (INDI) method, both controllers achieved highly aggressive trajectories within feasible limits.

The INDI method is commonly used in various controllers to improve output efficiency, managing nonlinearities and disturbances during large-angle operations while reducing sensitivity to model mismatch \cite{yang2023indi}. With the enhanced INDI method, MAVs gain precise control and maneuverability, facilitating navigation of complex flight paths and agile maneuvers \cite{zhou2021extended, ho2023incremental}.

Line-following control entails a robotic system's ability to detect and accurately track a predefined path or landmark \cite{gaspar2000vision}. This control method uses sensors like cameras or infrared sensors to capture visual data from the path. The system processes this data and generates control signals for steering the robot or UAV in alignment with the path. The control algorithm continuously calculates necessary adjustments, utilizing feedback mechanisms and Proportional Integral Derivative (PID) control techniques \cite{binugroho2015control} \cite{maniha2011two}. Due to the limited computational power, line followers typically avoid expensive line detection algorithms and hence reduce the computational cost \cite{mahaleh2018harmony}.

\section{Line Detection and Following}
\label{sec:method}
The focus of this study is to address the challenges of localizing and navigating MAVs in environments where GPS signals are not accessible. To overcome this limitation, a vision-based line-following algorithm is employed. The algorithm enables the MAV to actively determine its own position by analyzing the received data from its camera. These data include line detection algorithms, which are then fused together using a Kalman filter, leveraged from our previous work \cite{ho2013automatic}, to estimate the state of the target according to previous information captured by the camera. 

Once the MAVs position is estimated, it is utilized by the flight control unit to guide the MAVs navigation. This integration of vision-based localization and flight control allows the MAV to autonomously follow the designated flight path. To validate the effectiveness of the algorithm, simulations have been conducted using the IRIS copter in the Gazebo 11 environment, providing a realistic virtual platform for testing, and evaluating the performance of the MAV localization and navigation capabilities.

\subsection{Vision Based Detection}
This project utilizes vision-based detection, which involves using cameras or visual sensors to identify and analyze objects or patterns in an environment. It captures visual data and processes it using computer vision algorithms to extract valuable information, relying on visual cues like color, shape, texture, and motion to detect and classify objects.

Vision-based detection offers advantages like rich information in visual data, and real-time analysis for object recognition, tracking, and localization. It enables systems to understand and interact with their surroundings effectively. Advancements in algorithms, deep learning, and camera sensors continuously enhance this technology, opening new possibilities for visual understanding applications. After collecting image or video data, the color detection algorithm processes it. This algorithm identifies pixels corresponding to a specific color, turning them white (binary 1), while the rest becomes black (binary 0). This simplifies the image for further processing.

\begin{figure}[ht]
\centering
\includegraphics[trim=0 0 0 0, clip, width =0.4\textwidth]{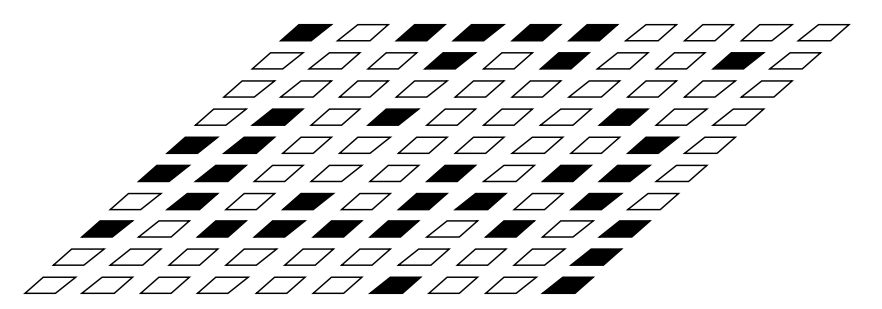}
\caption{A $10 \times 10$ binary image which is shown in black and white. \cite{bovik2009basic}}
\label{fig:binary}
\end{figure}

HSV color detection is used to identify and extract specific colors from a camera's video stream based on Hue, Saturation, and Value (HSV) components. Unlike RGB, which uses red, green, and blue channels, HSV separates colors into three components for intuitive color-based analysis. In HSV, Hue represents color (0-360 degrees), Saturation measures intensity, and Value indicates brightness. Lower and upper HSV ranges are set to capture a broader range of colors due to environmental factors. For example, the HSV values for yellow could be [18, 94, 140] to [48, 255, 255], with adjustments as needed. This method sets acceptable HSV value ranges to detect desired colors. Pixels in an image or video are compared to these thresholds, determining if they belong to the desired color range. This approach enhances MAV localization and tracking by isolating specific colors accurately, offering flexibility in object identification based on color characteristics.

\begin{figure}[ht]
\centering
\includegraphics[trim=0 0 0 0, clip, width =0.48\textwidth]{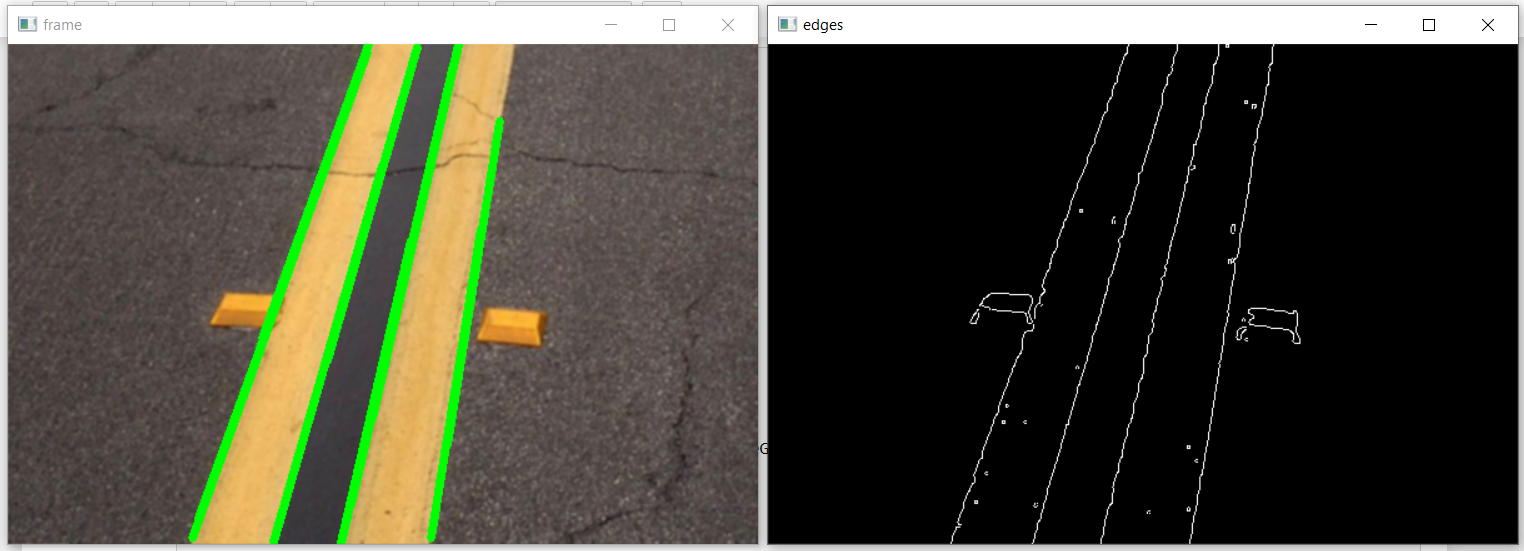}
\caption{HSV color detection on yellow line image.}
\label{fig:hsv}
\end{figure}

The algorithm has a drawback, as it can introduce noise into the image, affecting the control system. To enhance the output, morphological operations are applied to eliminate noise. Morphology in image processing aims to remove various forms of noise and textures while preserving the shapes and sizes of larger objects. The key processes are erosion and dilation. Dilation sets the output pixel's value to the maximum, turning it to 1 in binary if neighboring pixels are also 1, making lines thicker and objects more prominent. Erosion, on the other hand, sets the output pixel to 0 in binary, eliminating narrow lines and floating pixels. Once noise is removed, edge detection locates boundaries in the video stream by analyzing intensity variations. It identifies areas with significant pixel value changes, often representing object or region boundaries. The Canny edge detector is used in this study, among other techniques.

The Hough lines transform is employed to detect lines, even when parts are missing. This is valuable for scenarios like dashed road lines or partially obscured lines. The maximum line gap parameter is set as 50 to allow for the inclusion of line segments that have gaps between them. Line segments within this distance will be connected and considered as a single line. The centroid in the x-direction (cx) and y-direction (cy) is computed by finding the average number of accumulated detected lines. 

\subsection{Detection using Kalman Filter}

The Kalman filter is a mathematical algorithm used for estimating the state of a system by combining measurements from a camera view with predictions from a mathematical model. It provides an optimal estimation of the system's state, even in the presence of noise or uncertain measurements. The basic concept behind the Kalman filter is to iteratively update the estimated state of a system by combining a prediction based on the system dynamics and an observation obtained from camera measurements. The filter considers the uncertainty associated with both the predictions and the measurements to compute the optimal estimate of the system state. When it comes to line detection, the Kalman filter is used to track the position and trajectory of a line in the video sequence. The working principle of the Kalman filter is shown mathematically below.

To estimate the state of discrete-time controlled process,
\begin{equation}\label{e1}
    x_{k+1}=A_kx_k+Bu_k+w_k
\end{equation}

with the measurement of
\begin{equation}\label{e2}
    z_k=H_kx_k+v_k
\end{equation}

where $w_k$ and $v_k$ are the process noise and measurement noise, and they are independent. $A_k$ is an $n\times n$ matrix that relates the state at time step $k$ to the time step $k+1$, and $B$ is a $n \times 1$ matrix that relates to control input and the state $x$. Matrix $H$ is in $m\times n$ which relates to the state of measurement $z_k$.

Next, a priori estimate error covariance is defined as
\begin{equation}\label{e3}
    P^{-}_k=E[e^{-}_ke^{-T}_k]
\end{equation}

and posteriori estimate error covariance is
\begin{equation}\label{e4}
    P_k=E[e_ke^{T}_k]
\end{equation}

where $e_k^{-}$ and $e_k$ are priori and posteriori estimate errors respectively. The combination of priori estimate, $x_k^{-}$, and measurement innovation, which is the difference between the actual measurement, $z_k$, and measurement prediction, $H_kx_k^{-}$ is shown in (\ref{e5}).

\begin{equation}\label{e5}
    \hat{x}_k=\hat{x}_k^{-}+K\left(z_k-H_k \hat{x}_k^{-}\right)
\end{equation}

$K$ is an $n \times m$ matrix and it is the Kalman gain that minimizes posteriori error covariance. One of the ways of $K$ minimizes posteriori error is shown in (\ref{e6}.

\begin{equation}\label{e6}
    K_k=P_k^- H_k^T (H_k P_k^- H_k^T+R_k)^{-1}
\end{equation}

The final equations for time and measurement updates are shown below:

Time update
\begin{equation}\label{e7}
    \hat{x}_{k+1}^{-}=A_k \hat{x}_k+B u_k
\end{equation}
\begin{equation}\label{e8}
    P_{k+1}^{-}=A_k P_k A_k^T+Q_k
\end{equation}

Measurement update
\begin{equation}\label{e9}
    K_k=P_k^{-} H_k^T\left(H_k P_k^{-} H_k^T+R_k\right)^{-1} 
\end{equation}
\begin{equation}\label{e10}
    \hat{x}_k=\hat{x}_k^{-}+K\left(z_k-H_k \hat{x}_k^{-}\right) 
\end{equation}
\begin{equation}\label{e11}
    P_k=\left(I-K_k H_k\right) P_k^{-}
\end{equation}

Equations (\ref{e7}) and (\ref{e8}) show the state estimate and covariance estimate from time step $k$ to time step $k+1$. The Kalman gain, $K_k$ (\ref{e9}) needs to be computed first during the measurement update process. Next, obtain $z_k$ in equation (\ref{e2}) and generate posteriori sate estimate as in equation (\ref{e10}). The final step will be obtaining a posteriori error covariance estimate as in equation (\ref{e11}). Following each update of time and measurement, the Kalman Filter proceeds by iteratively using the previous posteriori estimates to predict the new priori estimates. This iterative process ensures that the filter continuously refines its estimates based on the most recent information available. Fig.~\ref{fig:kalmanchart} below shows the complete operation of the filter.

\begin{figure}[ht]
\centering
\includegraphics[trim=0 0 0 0, clip, width =0.48\textwidth]{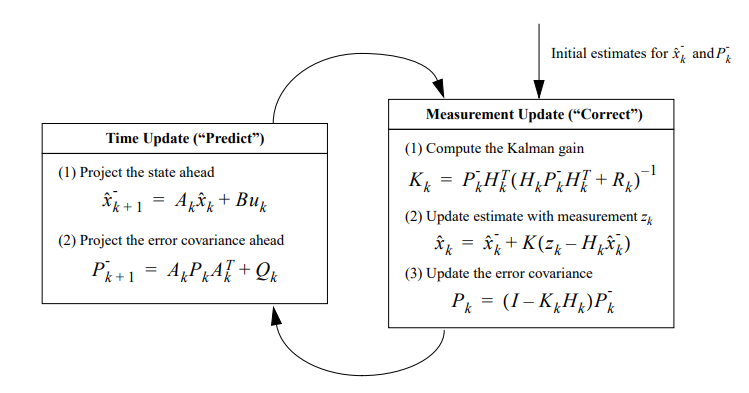}
\caption{The Kalman filter algorithm used with the line detection algorithm for the centroid estimation. \cite{bishop2001introduction}}
\label{fig:kalmanchart}
\end{figure}

\subsection{Autonomous Navigation}

This project focuses on utilizing two main classes for the line-following task with an MAV. The first class is responsible for line detection, while the second class handles the MAV navigation.

In the line detection class, an image is captured from the MAV onboard camera in the simulated environment. The class executes color thresholding, masking, and centroid detection to get the appropriate direction for the MAV to navigate. The direction is determined by evaluating the relative position of the centroid of the identified line with respect to the MAV current position. Based on this analysis, the class publishes corresponding commands such as moving forward, yawing left, yawing right, or yaw to search lines in cases where no line is detected.

The navigation class receives the direction commands published by the detection class. By utilizing the information, the class publishes velocity commands to the MAV, enabling it to effectively track the line and maintain its trajectory within the environment. Upon reaching the end of the path, the MAV initiates a search maneuver by hovering in place, actively seeking another line to follow.

\begin{figure*}[ht]
\centering
\includegraphics[trim=0 0 0 0, clip, width =0.7\textwidth]{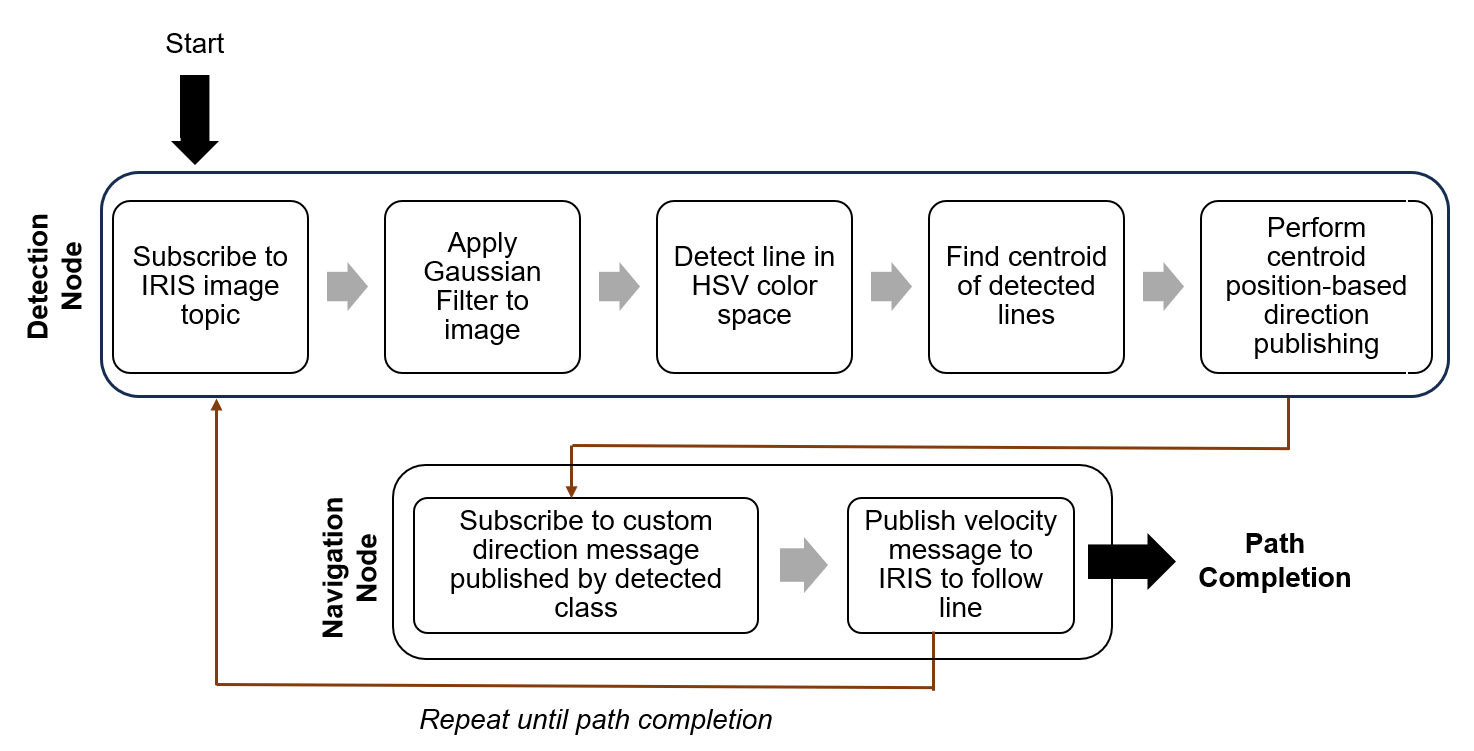}
\caption{An overview of the workflow of the autonomous line-following MAV between the detection nodes and navigation nodes.}
\label{fig:autonomous}
\end{figure*}

\section{Simulation}
\label{sec:Sim}
To evaluate the effectiveness of the developed algorithm, a dynamic model was simulated using the Gazebo platform. This simulation aimed to replicate the real flight conditions of a MAV within a virtual environment. By implementing the algorithm in the simulation, the behavior of the model was controlled and tested extensively, eliminating the need for a physical MAV. The use of the simulation provided several advantages during the algorithm development process. The STIL simulation method proved to be highly suitable for testing purposes. It allowed for comprehensive evaluations while safeguarding the physical MAV platform from any potential damage. Furthermore, the SITL simulation was not affected by weather conditions, enabling testing to be conducted at any time.

By leveraging the capabilities of the Gazebo platform and the SITL simulation environment, the algorithm could be thoroughly assessed, refined, and optimized without the constraints and risks associated with physical flight tests. This approach significantly contributed to the algorithm's development and ensured its reliability and effectiveness in real-world MAV operations.

Several simulations are carried out in different indoor environments through the Gazebo platform with the aid of ROS Noetic. The laptop utilized for running simulations is equipped with an Intel(R) Core(TM) i5-9300H CPU clocked at 2.40GHz, 8.00 GB of RAM, and features Intel(R) UHD Graphics 630. The first environment is a path of straight line with some corners and edges while the second tested environment is a circular path with curved turns. 

\subsection{Environment Setup}
\begin{figure}[ht]
\centering
\includegraphics[trim=0 0 0 0, clip, width =0.48\textwidth]{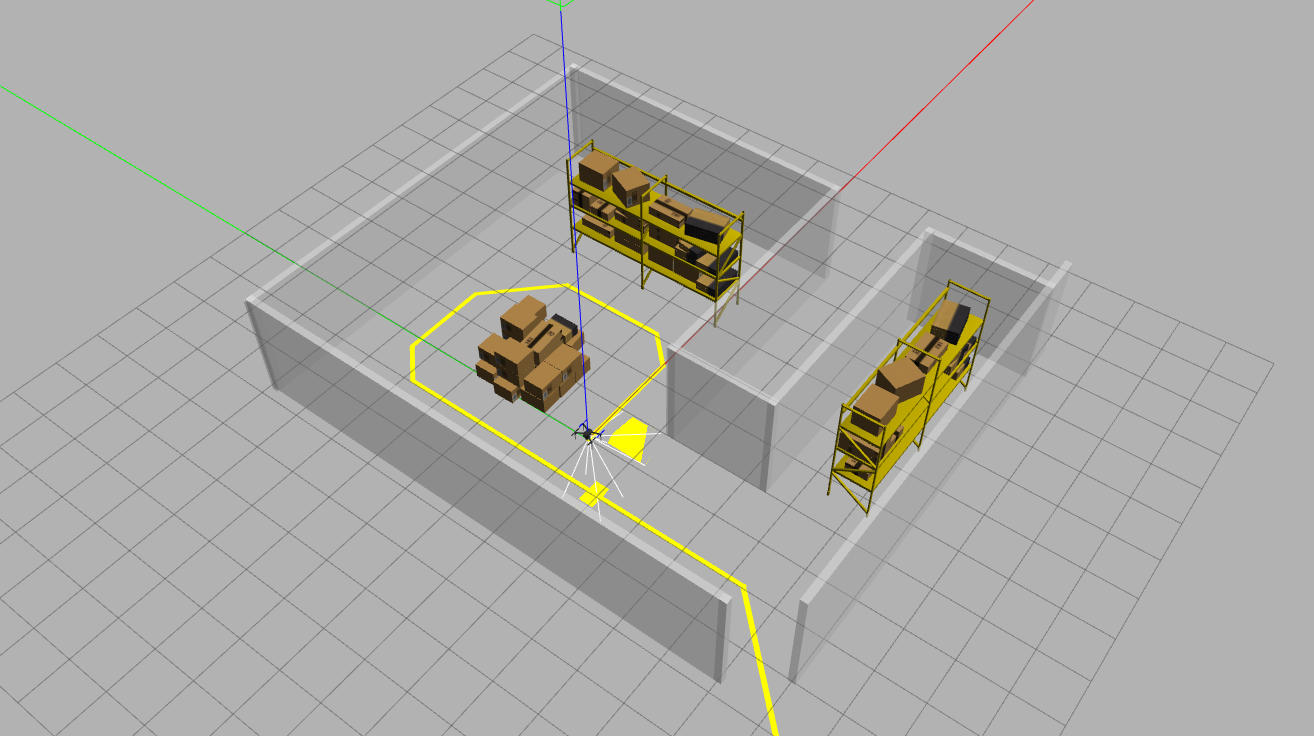}
\caption{Simulation Environment 1: A warehouse environment with a straight-lines path.}
\label{fig:environment1}
\end{figure}

The first environment to be tested is shown in Fig.~\ref{fig:environment1}\footnote{Modified from \url{https://github.com/sudrag/line_follower_turtlebot}}. It is a simple warehouse environment built up in a small space, equipped with two racks and a cluttering of boxes. The warehouse environment comprises a designated path that consists of a straight line with occasional corners. The straight-line sections allow for streamlined movement, while the corners introduce essential elements of navigation and pose specific challenges for the MAV control system.

\begin{figure}[ht]
\centering
\includegraphics[trim=0 0 0 0, clip, width =0.48\textwidth]{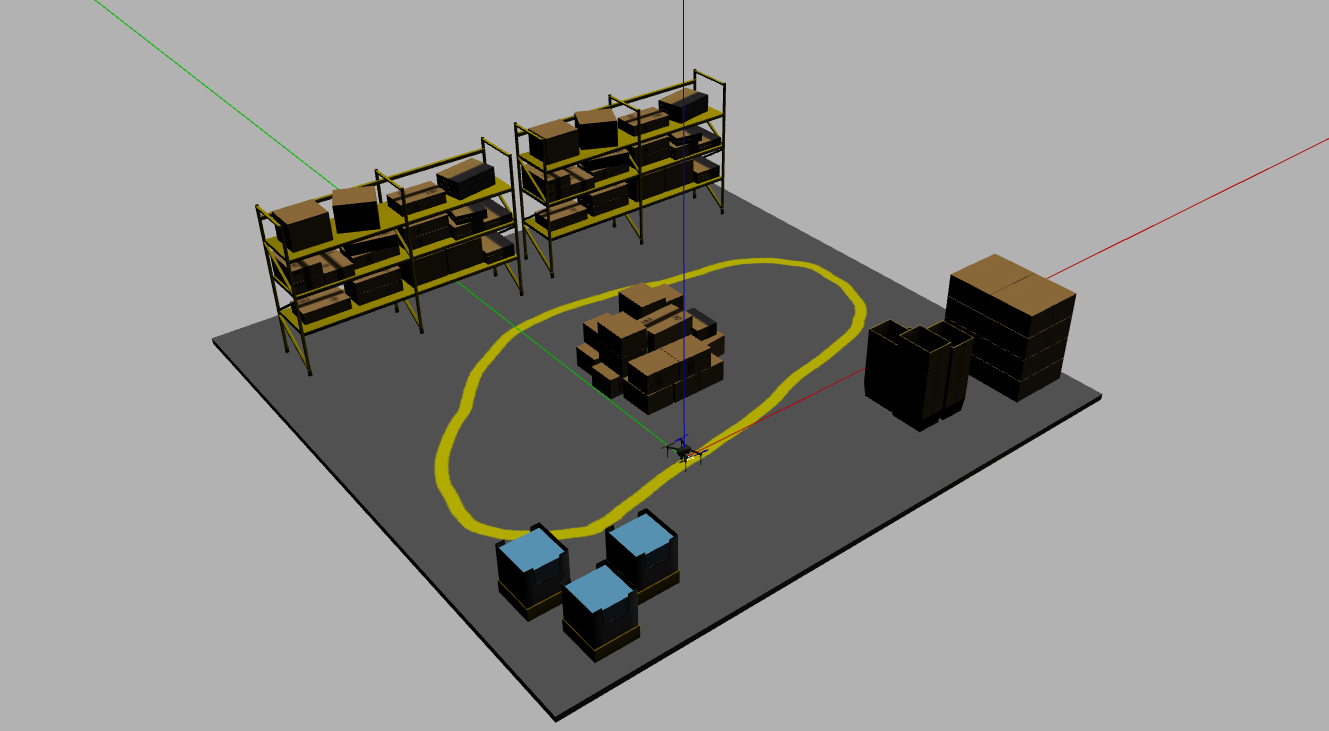}
\caption{Simulation Environment 2: A warehouse environment with a circular path.}
\label{fig:environment2}
\end{figure}

The second simulation environment to be tested is shown in Fig.~\ref{fig:environment2}\footnote{Modified from \url{https://github.com/DougUOW/line_follower_pkg/tree/master}}. It is a simple warehouse environment but with a circular path. Arcs can have varying radii, and the curvature of the line changes as the MAV moves. This requires the MAV to constantly adapt its turning radius and adjust its yawing angle to accurately follow the curve. It becomes more challenging for the MAV control system to maintain a precise trajectory while accounting for the changes.

\subsection{Performance Evaluation}
\begin{figure}[ht]
\centering
\includegraphics[trim=0 0 0 0, clip, width =0.48\textwidth]{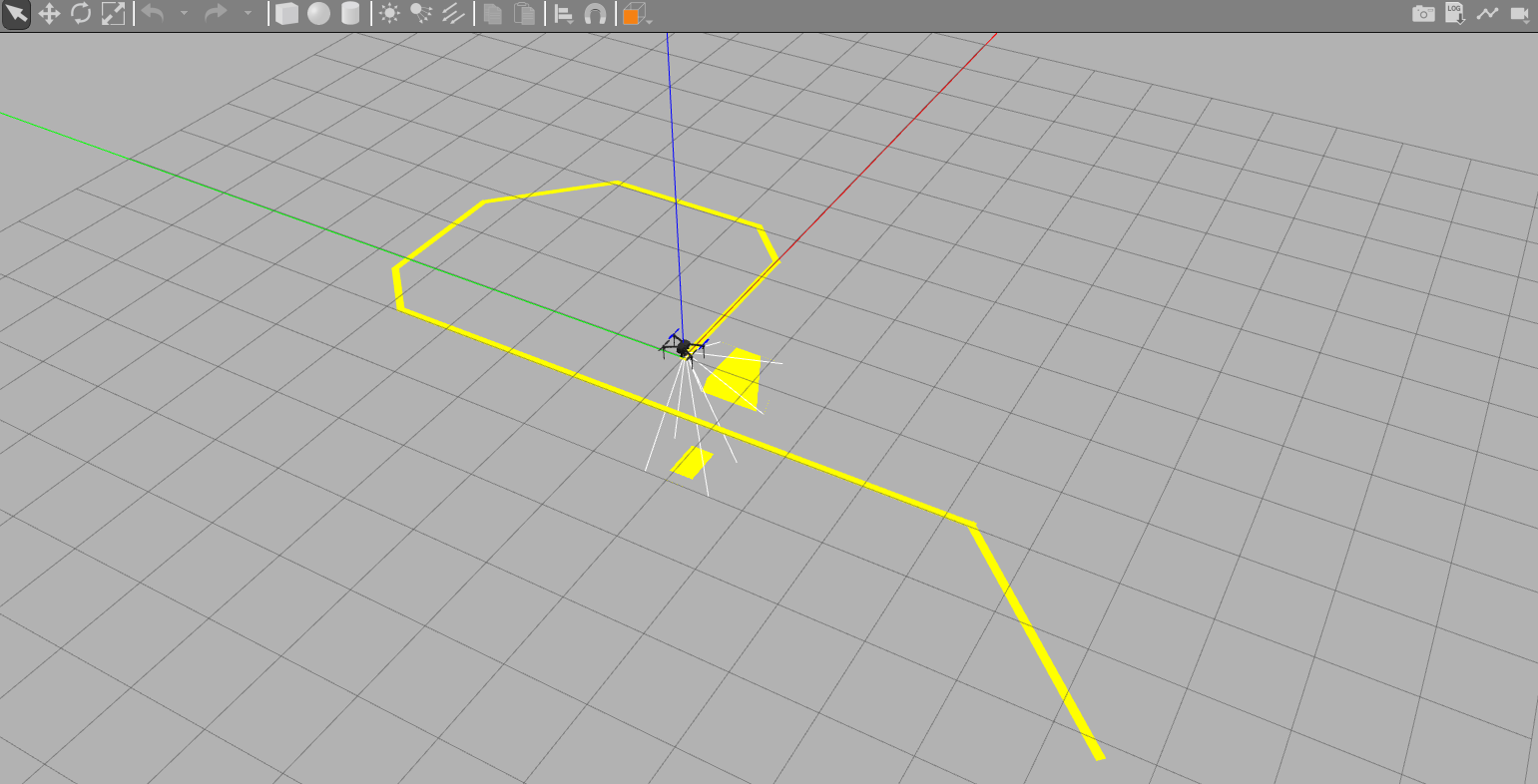}
\caption{A simple yellow line environment in the Gazebo platform was used to test the robustness of the line detection algorithm.}
\label{fig:simple_yellow}
\end{figure}

In order to get a more robust performance and result, the evaluation for the line detection algorithm is tested before running the SITL simulation in warehouse environments. Measurements such as HSV value and frame rate per second (FPS) are the methods chosen to test the performance of the Hough Line Transform HSV color detection employed in this study. The measurements are carried out in a simple yellow path environment as shown in Fig.~\ref{fig:simple_yellow}.

\begin{figure}[ht]
\centering
\includegraphics[trim=0 0 0 0, clip, width =0.48\textwidth]{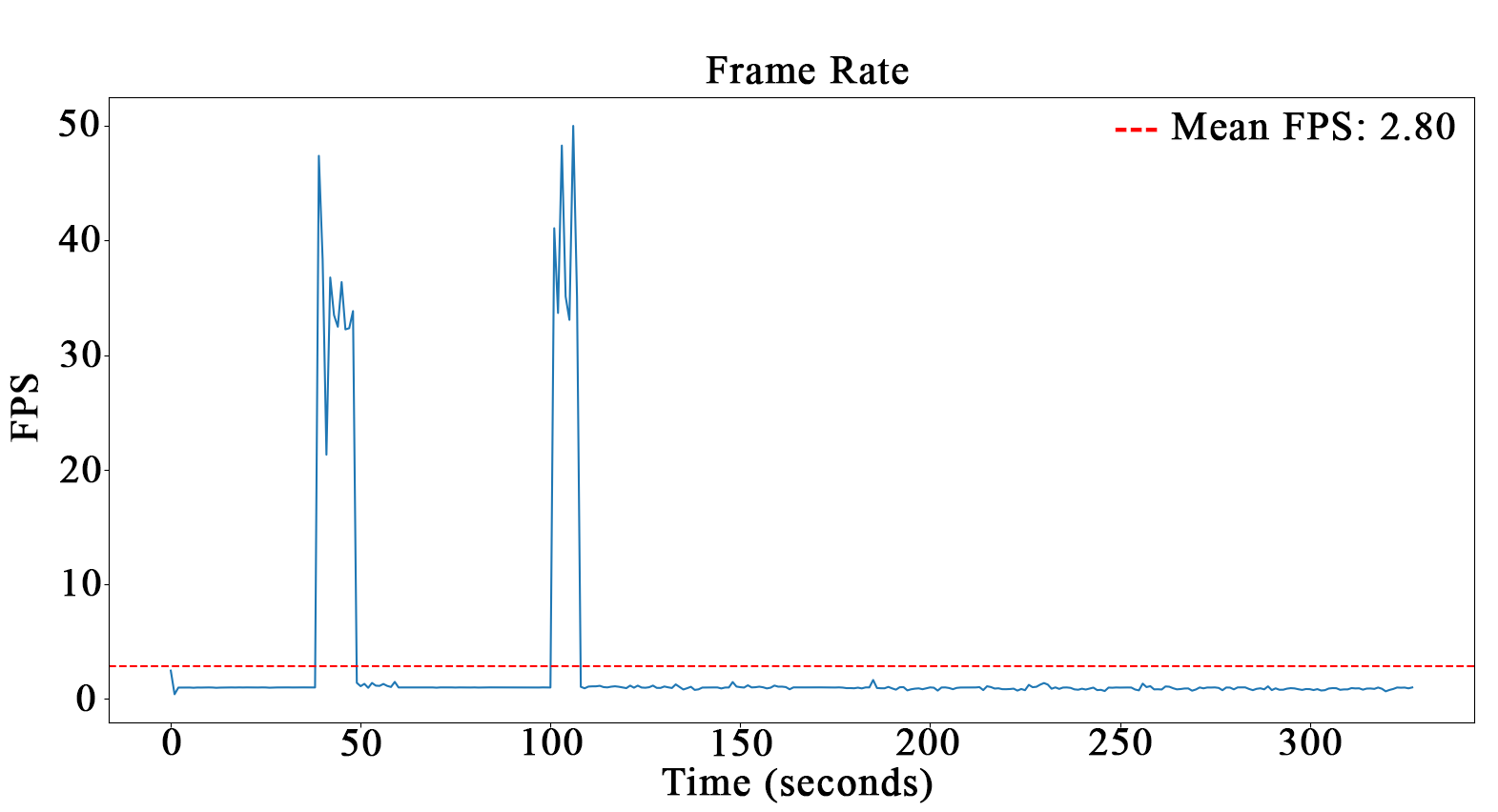}
\caption{Frame rates measured when running the Gazebo simulation.}
\label{fig:fps}
\end{figure}

Frame rate per second (FPS) is a measure of how many individual frames or images are displayed or captured in one second of time. It is used to describe the smoothness and fluidity of motion in video or animation. In the context of line detection algorithms, FPS plays a crucial role in real-time applications where high frame rates are desired for accurate and timely results. The frame rate of a line detection algorithm determines how quickly it can process incoming frames and produce line detection results. Higher frame rates allow for smoother and more responsive processing, which is particularly important in real-time applications where timely decisions or actions are required. The FPS result of the algorithm used is shown in Fig. ~\ref{fig:fps}. 

The mean FPS of the Gazebo platform during simulation is only 2.80 with a maximum of 50.02 and a minimum of 0.38. The FPS normally ranges between 0.8-1.0 but it fluctuated vigorously over a certain duration. The low FPS and the unstable are due to factors such as hardware limitations and software inefficiencies. Insufficient CPU and GPU power lead to the result of low FPS when dealing with the line detection algorithm which is a computationally intensive task. Next, the low-end graphics card struggles to handle demanding visual computations, hence impacting the FPS. Other than that, the poor memory allocation also results in frequent memory access delays, impacting the processing speed. Processing high-resolution frames and working with large datasets also significantly increases the computational load and slows down the FPS. 

The altitude of the MAV plays a crucial role in the performance of a line-following algorithm. As the MAV altitude changes, it affects various aspects of the line following capabilities and overall flight behavior. Firstly, altitude directly influences the perspective and field of view (FOV) of the onboard camera used for line detection. Different altitudes can result in variations in the size, position, and clarity of the detected line. Secondly, altitude affects the MAV proximity to the line being followed. As the MAV flies higher or lower, the distance between the MAV and the line changes. This distance alteration affects the accuracy and responsiveness of the MAV control system. Additionally, changes in altitude affect the environmental conditions experienced by the MAV.

\begin{figure}[ht]
\centering
\includegraphics[trim=0 0 0 0, clip, width =0.48\textwidth]{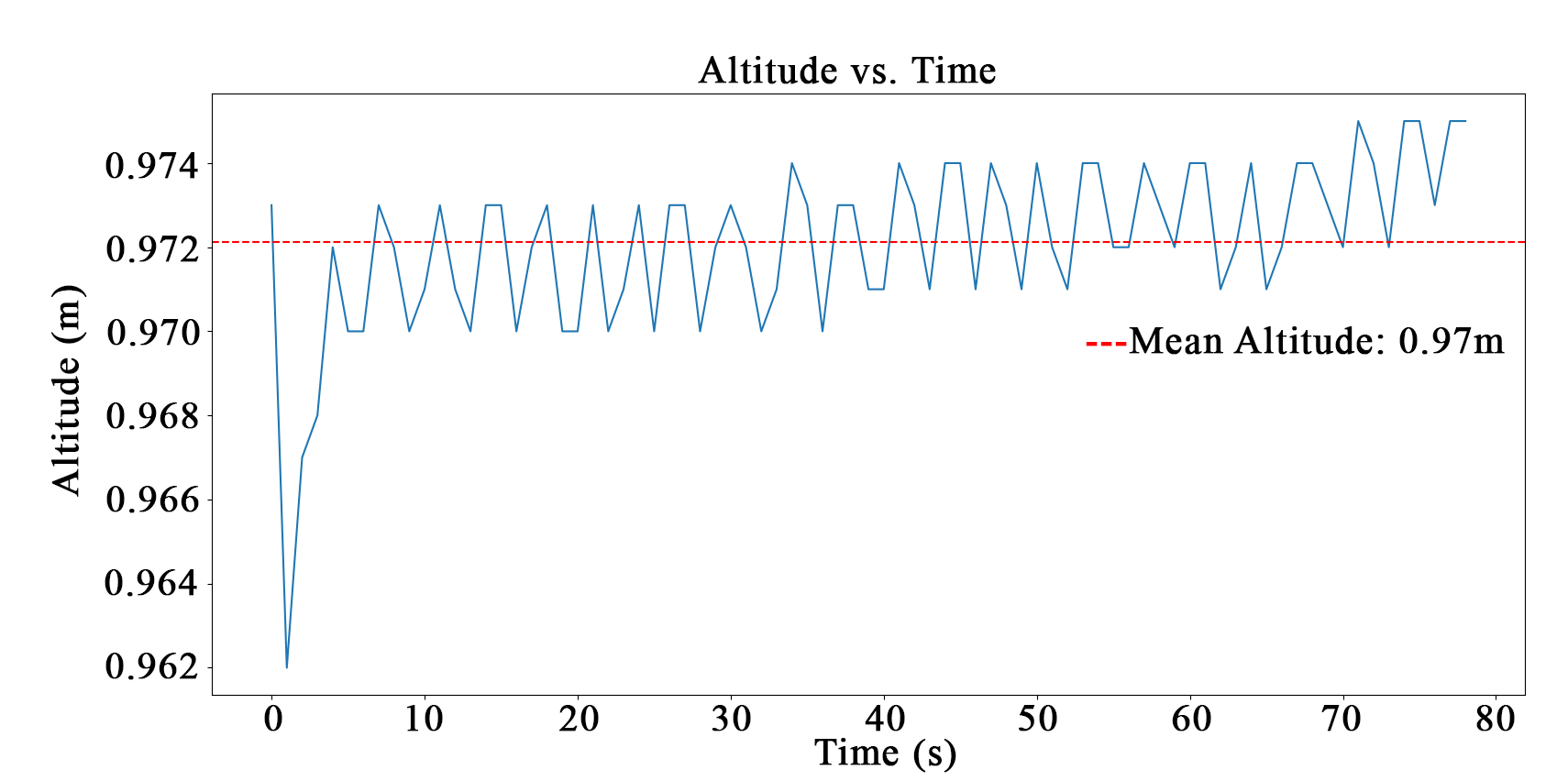}
\caption{Altitude measurements of the MAV along the flight path.}
\label{fig:altitude}
\end{figure}

The altitude of MAV from takeoff to landing is shown in Fig. ~\ref{fig:altitude}. The target altitude was set to be 1~meter. Based on the plot, the altitude of MAV ranges from 0.962m to 0.975m with a mean altitude of 0.97m. Overall, the altitude of the MAV was maintained at a stable range and did not fluctuate violently. This can help to maintain the line detection proximity and hence get a robust control responsiveness.

\begin{figure}[ht]
\centering
\includegraphics[trim=0 0 0 0, clip, width =0.25\textwidth]{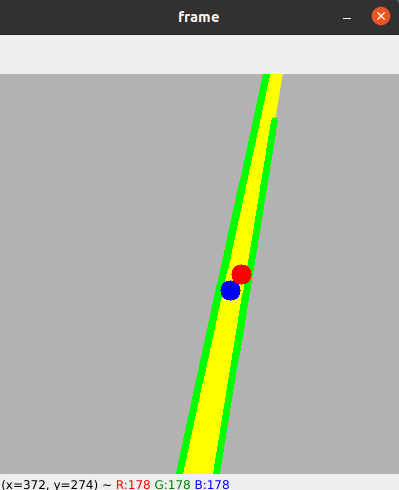}
\caption{Centroids computed from the line detection algorithm. The blue dot represents the original centroid, while the red dot represents the centroid estimated using the Kalman filter.}
\label{fig:kalman_frame}
\end{figure}

A Kalman filter is used in this work to predict and measure the centroid of the line detected through the onboard camera. The line detection system obtains measurements of the line's position, which is the centroid coordinates from camera vision. Then, the Kalman filter will predict the next state of the centroid based on the system dynamics model. This prediction is made by using the previous state estimate and the dynamics equations that govern the centroid's motion. Next, it compares the predicted state with the actual measurement obtained initially by updating the estimated state and covariance matrix using the Kalman gain. Finally, the estimated centroid positions are obtained from the Kalman filter as the output and the result is shown in Fig.~\ref{fig:kalman_frame}. 

\begin{figure}[ht]
\centering
\includegraphics[trim=0 0 0 0, clip, width =0.5\textwidth]{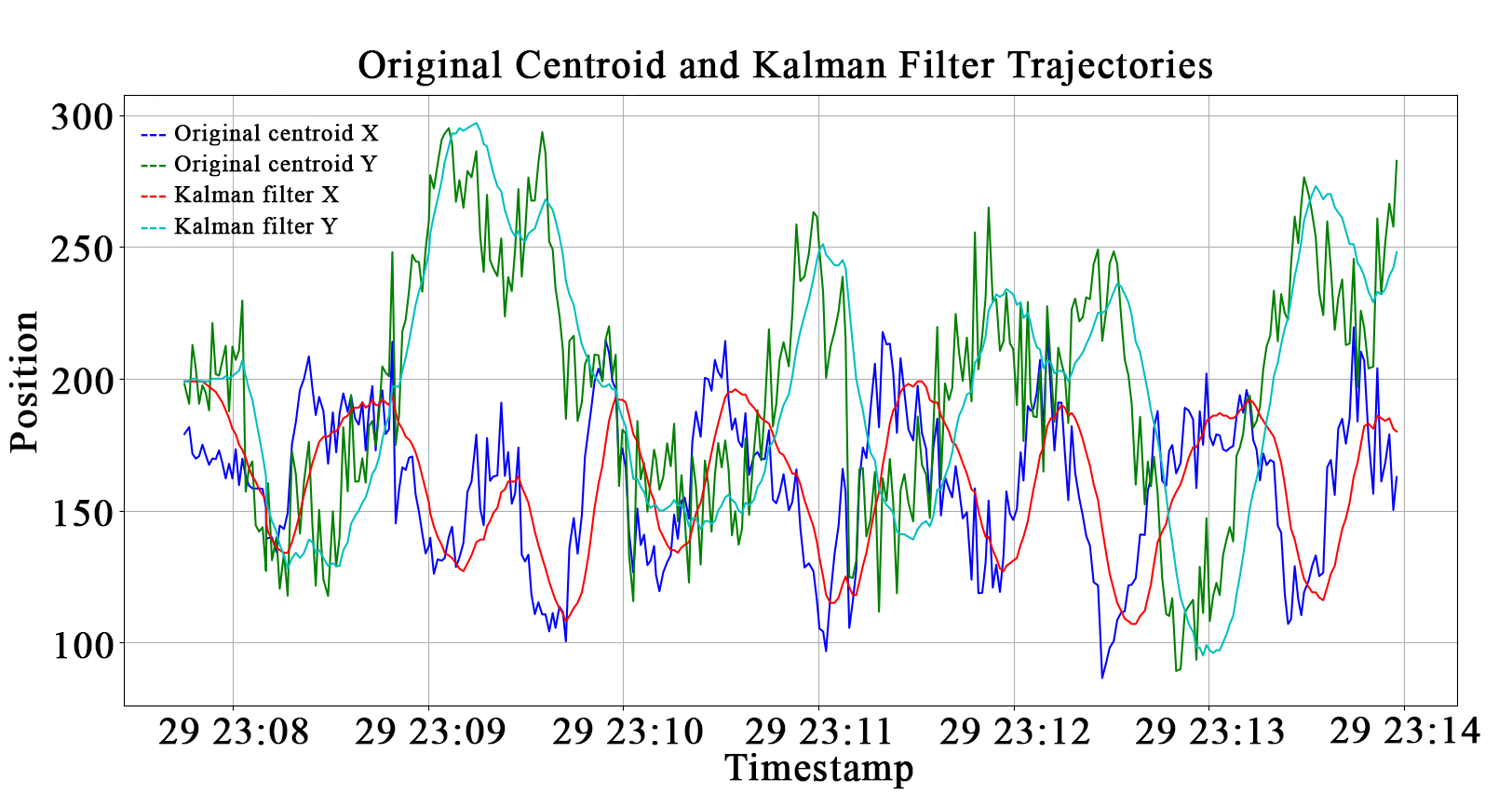}
\caption{The comparison between the original centroid and centroid estimated using the Kalman filter.}
\label{fig:kalman}
\end{figure}

According to Fig.~\ref{fig:kalman}, the trajectory of the original centroid position might appear more erratic and noise. The lack of filter results in a scattered plot, where the centroid position jumps around, especially in the presence of noise or sudden changes. On the other hand, with the application of a Kalman filter, the plotted centroid positions exhibit smoother and more consistent behavior over time. The Kalman filter helps in reducing the effects of noise and provides a more accurate estimate of the centroid's position. Consequently, the plotted points will show less variation and a more coherent trajectory. By comparing the plots side by side, it shows that the plot with a Kalman filter will exhibit smoother and more stable centroid positions. The plot without a filter, on the other hand, displayed more erratic movements and larger deviations due to the influence of noise. Overall, the Kalman filter improves the accuracy and stability of line centroid detection by dynamically adjusting the estimated position based on the measurements and the system's behavior, leading to more reliable line tracking and control.

\subsection{Flight Tests}

\begin{figure}[ht]
\centering
\includegraphics[trim=0 0 0 0, clip, width =0.40\textwidth]{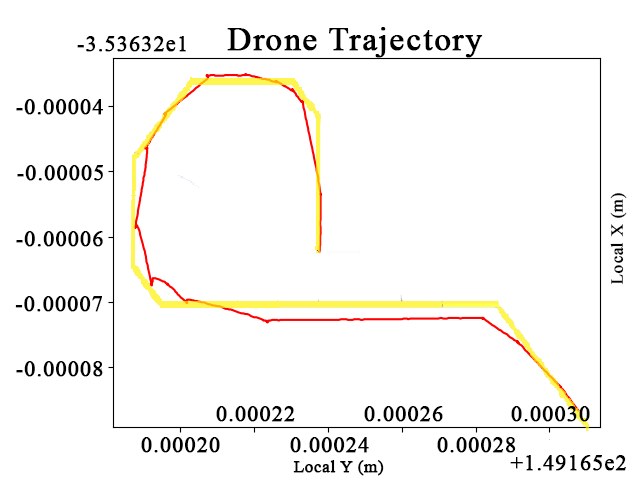}
\caption{2D flight trajectory of the MAV compared with the pre-defined path in Simulation Environment 1.}
\label{fig:T1}
\end{figure}

\begin{figure}[ht]
\centering
\includegraphics[trim=0 0 0 0, clip, width =0.48\textwidth]{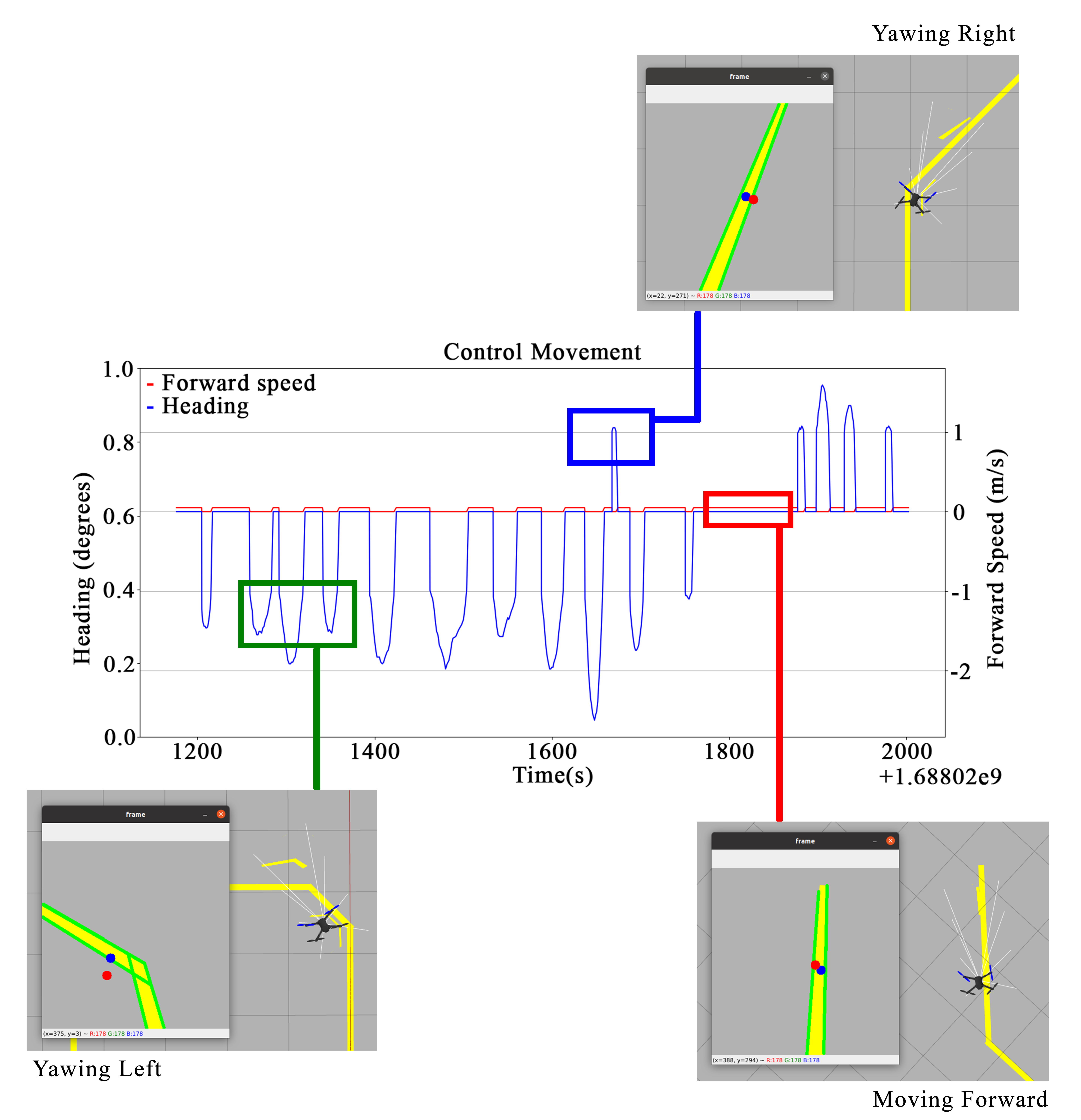}
\caption{The heading and forward speed of the MAV over time in Simulation Environment 1. When the MAV yaws to the left, the yaw angle will be negative, and the forward speed is zero. When the MAV moves forward, the forward speed remains constant, while when the MAV yaws to the right, the yaw angle is positive, and the forward speed is zero.}
\label{fig:H1}
\end{figure}

Fig.~\ref{fig:T1} illustrates the real flight trajectory (red) executed in the local X and local Y plane, comparing it to the pre-defined flight path (yellow) in Environment 1. The MAV successfully follows the pre-defined flight route, displaying the feasibility of the HSV line detection. However, some positional errors, particularly at the turning corners are observed, which can be attributed to the low FPS of the feedback controller. In Fig.~\ref{fig:H1}, the heading and forward speed of the MAV are illustrated. The MAV maintains a constant speed of 0.05 m/s throughout the flight. As shown in the figure, when the Kalman state centroid is located in the middle of the detection frame, the MAV moves forward at a constant speed. However, when the centroid of the detected line shifts to the left side of the detection frame, the MAV stops receiving velocity input and yaws to the left, resulting in a negative degree value. Similarly, the yawing motion to the right follows a similar pattern but with a positive degree value.

\begin{figure}[ht]
\centering
\includegraphics[trim=0 0 0 0, clip, width =0.48\textwidth]{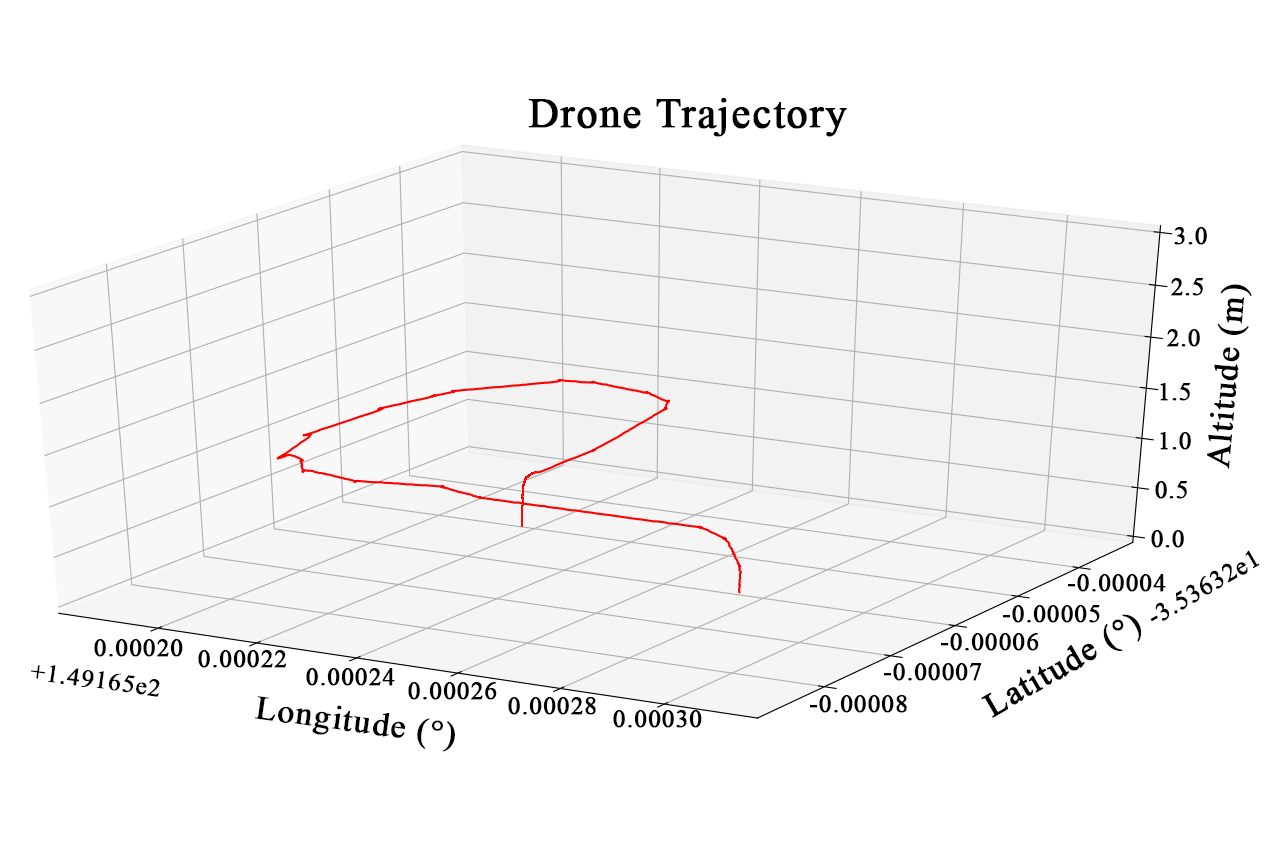}
\caption{3D flight trajectory of the MAV in Simulation Environment 1.}
\label{fig:3D1}
\end{figure}

\begin{figure}[ht]
\centering
\includegraphics[trim=0 0 0 0, clip, width =0.4\textwidth]{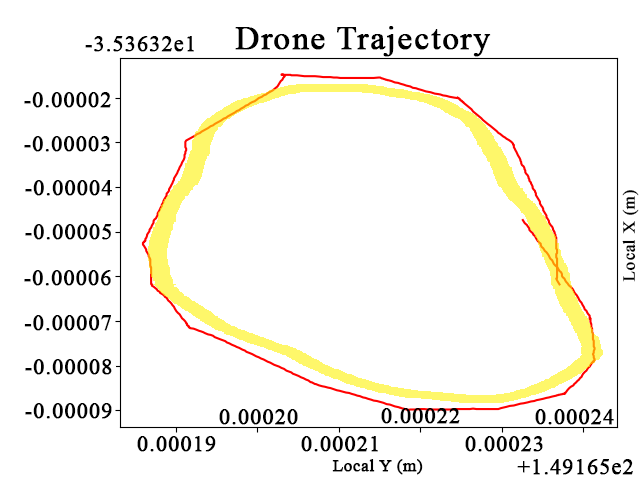}
\caption{2D flight trajectory of the MAV compared with the circular pre-defined path in Simulation Environment 2.}
\label{fig:T2}
\end{figure}

Fig.~\ref{fig:T2} illustrates the real flight trajectory (red) in the local X and local Y plane, comparing it to the pre-defined flight path (yellow) in Environment 2. Although the results indicate a delayed response in the yawing control command, causing the real flight trajectory to deviate slightly from the predetermined route, but the MAV successfully follows the pre-defined flight route. The delay and noise of the measurement can be analyzed using the flight data \cite{ho2016characterization} to improve the adaptability of the controller in the future. 

\begin{figure}[ht]
\centering
\includegraphics[trim=0 0 0 0, clip, width =0.48\textwidth]{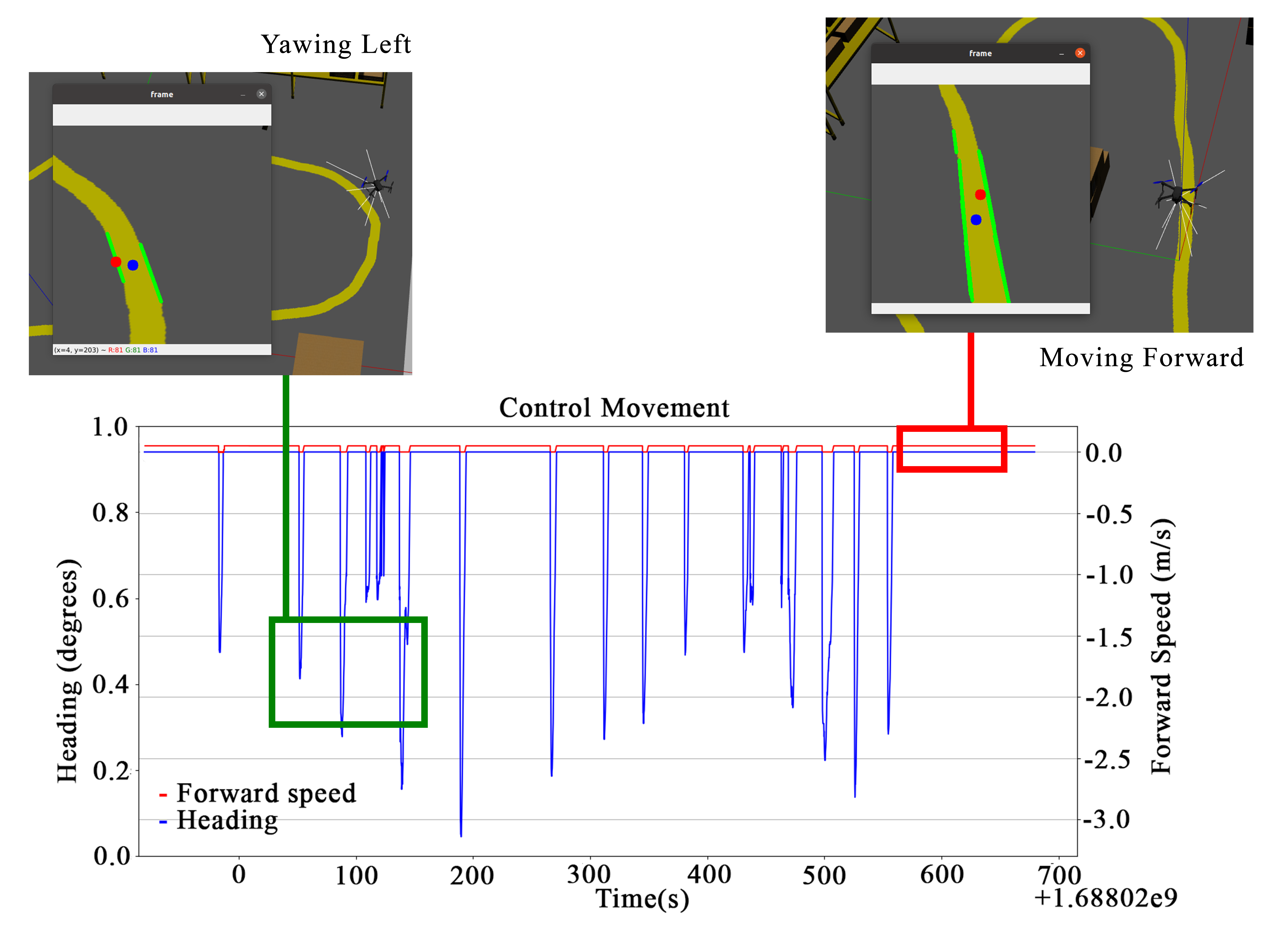}
\caption{The heading and forward speed of the MAV over time in Simulation Environment 2. When the centroid detected is in the center of the frame, the MAV receives forward velocity input but zero yaw input. The MAV receives negative yaw input and zero forward speed when the centroid is at the left side of the detection frame.}
\label{fig:H2}
\end{figure}

In Fig.~\ref{fig:H2}, the heading and forward speed of the MAV are presented. In comparison to the heading of the MAV in Environment 1, the heading of the MAV in Environment 2 demonstrates a more intense yaw angle due to the varying radii of the arc path. When the Kalman state centroid is situated in the middle of the detection frame, the MAV moves forward at a constant speed. However, when the centroid of the detected line shifts to the left side of the detection frame, the MAV stops receiving velocity input and yaws accordingly.

\begin{figure}[ht]
\centering
\includegraphics[trim=0 0 0 0, clip, width =0.48\textwidth]{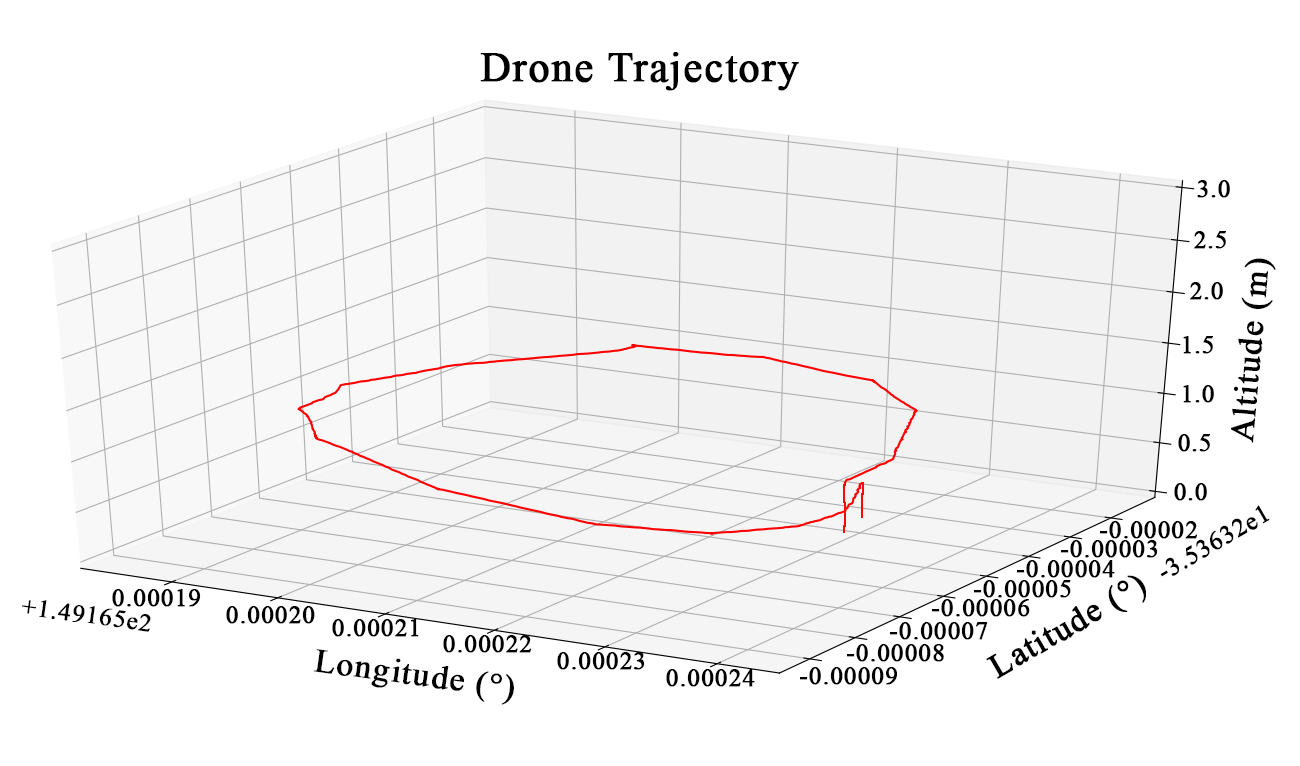}
\caption{3D flight trajectory of the MAV in Simulation Environment 2.}
\label{fig:3D2}
\end{figure}

\section{Conclusion}
\label{sec:concl}
This study introduces a vision-based control system that enables the localization of a MAV and autonomous navigation along paths. The aim of this integrated system is to facilitate the future development of MAV in autonomous warehouses. The system utilizes an onboard camera as a detector to recognize and determine the location of the line within an image. In this study, the HSV color detection algorithm was chosen for color detection. The reason behind this selection is its ability to handle variations in lighting conditions effectively, making it suitable for real-world environments. Moreover, the HSV color space offers a more intuitive and flexible representation of colors, which simplifies the thresholding and segmentation process for identifying the desired color range. The centroids of the detected lines are computed using an edge detection technique.

Furthermore, the integrated system introduced in this study incorporates the Kalman filter to enhance its performance. The Kalman filter is employed to predict the next state estimate by considering the previous state estimate and the system dynamics. Subsequently, the newly acquired centroid measurement is used to update the state estimate. By considering both the system dynamics and the noisy measurements, the Kalman filter aids in refining and smoothing the estimated centroid position. The updated centroid position, obtained through the Kalman filter, is then utilized for subsequent control actions within the line following system.

The effectiveness of the proposed system for autonomous line following was evaluated through simulation tests. These tests involved the MAV autonomously navigating in various environments with different pre-defined path shapes. The results of the simulation tests demonstrated that the HSV line detection used in the system is robust. The MAV successfully followed the pre-defined flight route, indicating the system's capability. Although there were some position errors observed at the turning points, the actual trajectory of the MAV demonstrated that it could navigate the entire pre-defined path accurately through the output of the vision algorithm.


%



\ifCLASSOPTIONcaptionsoff
  \newpage
\fi



%


\bibliographystyle{IEEEtran}
\bibliography{IEEEabrv,ref}


\vfill


\end{document}